\documentclass[acmtog]{acmart}
\AtBeginDocument{%
  }

\setcopyright{acmlicensed}
\copyrightyear{2025}
\acmYear{2025}
\acmDOI{XXXXXXX.XXXXXXX}
\acmConference[SA Conference Papers '25]{SIGGRAPH Asia 2025 Conference Ppaers}{December 15--18,
  2025}{Hong Kong, China}
\acmISBN{978-1-4503-XXXX-X/2018/06}

\acmSubmissionID{2563}

\usepackage[T1]{fontenc}
\usepackage{hyperref}
\usepackage{url}
\usepackage{graphicx}
\usepackage{booktabs}
\usepackage[most]{tcolorbox}
\usepackage{listings}
\usepackage{subcaption}
\usepackage{pifont}
\usepackage{array}
\usepackage{multirow}
\usepackage{makecell}

\usepackage{wrapfig,lipsum}
\usepackage[ruled,vlined]{algorithm2e}

\usepackage{pgfplots}
\pgfplotsset{compat=1.12} 
\usepackage{xcolor}
\usepackage{tikz}
\usepackage{xspace}
\usepackage{makecell}
\usepackage{soul}
\usepackage{amsmath}
\usepackage{subcaption}
\usepgfplotslibrary{groupplots}
\usetikzlibrary{angles,quotes} 
\usetikzlibrary{shapes,arrows}
\usetikzlibrary{backgrounds}
\usetikzlibrary{matrix}
\usetikzlibrary{patterns}
\usepackage{tikzsymbols}
\usepackage{tikz-3dplot}
\usepackage{tabularx}
\usepackage[utf8]{inputenc}

\definecolor{darkgreen}{rgb}{0.0, 0.8, 0.0}

\definecolor{navy}{RGB}{0, 0, 128}

\newcommand{\ours}{ComfyUI-R1}

\begin{document}

\title{ComfyUI-R1: Exploring Reasoning Models for Workflow Generation}


\author{Zhenran Xu}
\authornote{Both authors contributed equally to this research.}
\affiliation{%
  \institution{Harbin Institute of Technology (Shenzhen)}
  \city{Shenzhen}
  \country{China}}
\email{xuzhenran@stu.hit.edu.cn}

\author{Yiyu Wang}
\authornotemark[1]
\affiliation{%
  \institution{Alibaba International Digital Commerce}
  \city{Hangzhou}
  \country{China}}
\email{wangyiyu18@mails.ucas.ac.cn}

\author{Xue Yang}
\affiliation{%
  \institution{Alibaba International Digital Commerce}
  \city{Hangzhou}
  \country{China}}
\email{yx9966@126.com}

\author{Longyue Wang}
\affiliation{%
  \institution{Alibaba International Digital Commerce}
  \city{Hangzhou}
  \country{China}}
\email{vincentwang0229@gmail.com}

\author{Weihua Luo}
\affiliation{%
  \institution{Alibaba International Digital Commerce}
  \city{Hangzhou}
  \country{China}}
\email{weihua.luowh@alibaba-inc.com}

\author{Kaifu Zhang}
\affiliation{%
  \institution{Alibaba International Digital Commerce}
  \city{Hangzhou}
  \country{China}}
\email{kaifu.zkf@alibaba-inc.com}

\author{Baotian Hu}
\authornote{Corresponding author.}
\affiliation{%
  \institution{Harbin Institute of Technology (Shenzhen)}
  \city{Shenzhen}
  \country{China}}
\email{hubaotian@hit.edu.cn}

\author{Min Zhang}
\affiliation{%
  \institution{Harbin Institute of Technology (Shenzhen)}
  \city{Shenzhen}
  \country{China}}
\email{zhangmin2021@hit.edu.cn}


\begin{abstract}
AI-generated content has evolved from monolithic models to modular workflows, particularly on platforms like ComfyUI, enabling customization in creative pipelines. 
However, crafting effective workflows requires great expertise to orchestrate numerous specialized components, presenting a steep learning curve for users. 
To address this challenge, we introduce \textbf{ComfyUI-R1}, the first large reasoning model for automated workflow generation. 
Starting with our curated dataset of 4K workflows, we construct long chain-of-thought (CoT) reasoning data, including node selection, workflow planning, and code-level workflow representation. 
ComfyUI-R1 is trained through a two-stage framework:
(1) CoT fine-tuning for cold start, adapting models to the ComfyUI domain;
(2) reinforcement learning for incentivizing reasoning capability, guided by a fine-grained rule-metric hybrid reward, 
ensuring format validity, structural integrity, and node-level fidelity. 
Experiments show that our 7B-parameter model achieves a 97\% format validity rate, along with high pass rate, node-level and graph-level F1 scores, significantly surpassing prior state-of-the-art methods that employ leading closed-source models such as GPT-4o and Claude series. 
Further analysis highlights the critical role of the reasoning process and the advantage of transforming workflows into code.
Qualitative comparison reveals our strength in synthesizing intricate workflows with diverse nodes, 
underscoring the potential of long CoT reasoning in AI art creation.


\end{abstract}


\begin{teaserfigure}
\centering
  \includegraphics[width=\textwidth]{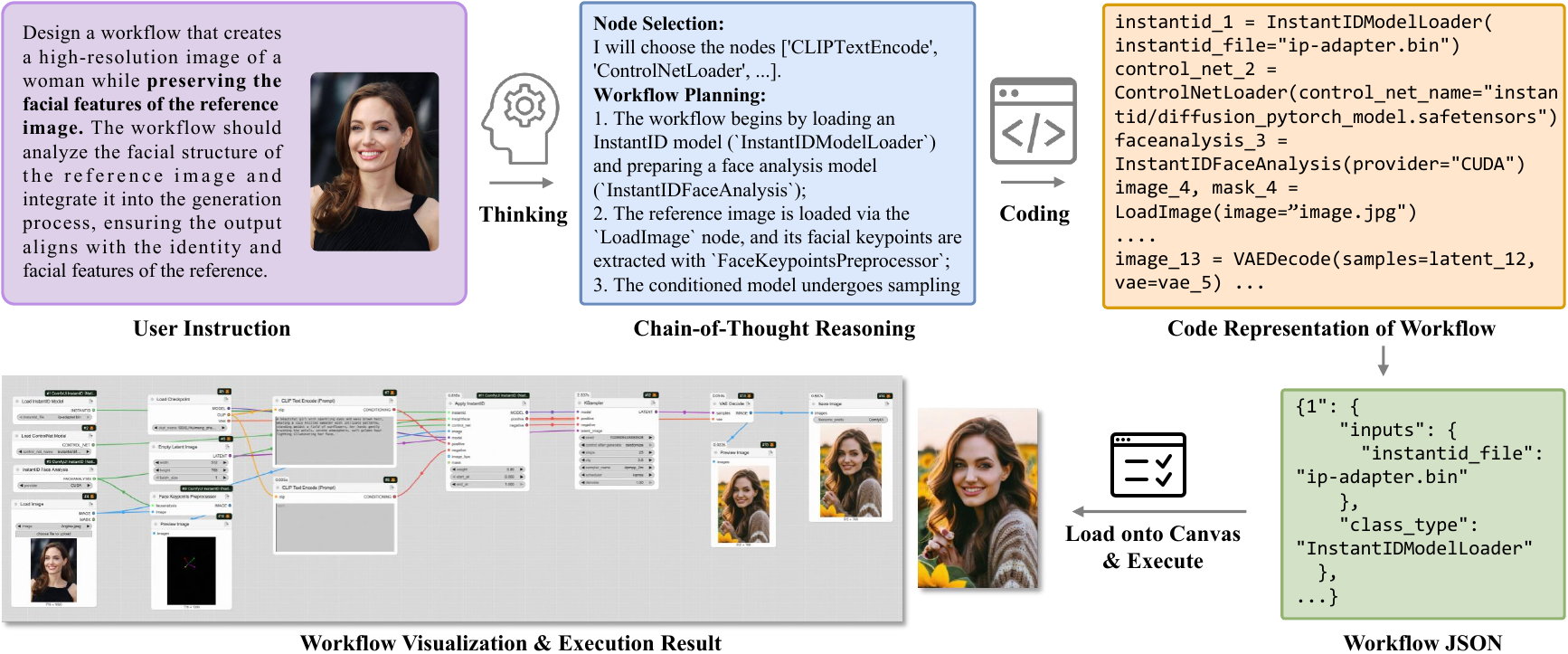}
  \caption{\textbf{We introduce ComfyUI-R1, a large reasoning model for automated workflow generation.} Given a user instruction, ComfyUI-R1 performs long chain-of-thought reasoning to generate a code representation of a ComfyUI workflow. The generated workflow adheres to the correct format, executes successfully, and produces an image that aligns with the user's instruction. ComfyUI-R1 is integrated in \textcolor{blue}{\url{https://github.com/AIDC-AI/ComfyUI-Copilot}}.}
  \label{fig:intro}
\end{teaserfigure}


\maketitle

\section{Introduction}

Recent advancements in large language models (LLMs) and image generation methods have democratized AI-generated content (AIGC) production~\cite{NEURIPS2020_4c5bcfec_diffusion,cyclediffusion,NEURIPS2021_49ad23d1_diffusion,anim-director}.
Instead of relying on an end-to-end diffusion model,
recent AIGC production has evolved into building advanced workflows to further enhance image quality~\cite{gal2024comfygen}.
Open-source frameworks such as ComfyUI~\cite{comfyanonymous2023comfyui} are emerging as important tools for low-code AI workflow development. 
Serving over 4 million active users and backed by a vibrant community contributing 12K components (e.g., SDXL~\cite{podell2023sdxl}, LoRA~\cite{ryu2023low}, ControlNet~\cite{zhang2023adding}, and IP-Adapter~\cite{ye2023ip}), 
ComfyUI enables flexible workflow orchestration via drag-and-drop components for multimodal tasks such as text-to-image generation, face swapping, and video editing.

Despite its convenient interactive interface, ComfyUI presents challenges for new users who may lack the knowledge and reasoning skills required to build effective workflows~\cite{huang2025comfygpt}. 
The large number of nodes, along with their interdependencies and complex configurations, demands extensive documentation knowledge to go through. 
Moreover, the intricate connections within workflows require strong planning abilities to coordinate modules with diverse functionalities. 
As a result, constructing a well-designed workflow often requires considerable expertise, 
making automated end-to-end workflow generation and user onboarding crucial for broader adoption.

Recently, there has been increasing research on automated ComfyUI workflow generation. 
These approaches leverage state-of-the-art LLMs (e.g, GPT-4o) to generate valid JSON-formatted workflows and break down the task into sub-tasks for multi-agent collaboration~\cite{xue2024comfybench,huang2025comfygpt,guo2025comfymind}. 
However, these methods still face several limitations: 
(1) Most studies focus only on text-to-image generation~\cite{sobania2024comfygi,gal2024comfygen}, limiting their applicability to more conditional image and video generation tasks; 
(2) The generated JSON files often fail to transform into structured workflows or contain hallucinated nodes, possibly due to GPT-4o's limited knowledge of newly introduced nodes and cumulative errors in multi-agent systems~\cite{cemri2025multiagentllmsystemsfail,wang-etal-2024-rethinking-bounds}. 
To address these issues, 
our work expands the scope to a wide range of multimodal tasks and focuses on post-training foundational LLMs rather than relying on commercial APIs.

The success of long chain-of-thought (CoT) reasoning has attracted widespread attention, exemplified by models such as OpenAI o1~\cite{openai2024introducing}, o3~\cite{openai2025o3o4}, and DeepSeek-R1~\cite{deepseekai2025deepseekr1}, which have achieved remarkable accuracy in solving Olympiad-level math problems.
Inspired by this progress, we introduce \textbf{ComfyUI-R1}, the first 
large reasoning model designed for automated workflow generation.
Specifically, we begin by collecting a substantial volume of workflows and node data from ComfyUI community websites, covering a diverse set of tasks including image, video, and 3D generation and editing. After a comprehensive cleaning process, we retain 4K workflows out of the original 27K.
Each example includes JSON-formatted and code representations of the workflow, along with a description of its functionality. 
Then, we generate long CoT reasoning sequences with a simulated node retrieval process, node selection, and workflow planning, ultimately producing the code representation of the workflow.

With the data prepared, ComfyUI-R1 is trained using a two-stage pipeline:
(1) Cold-start CoT fine-tuning to adapt models to the ComfyUI domain, addressing the \textit{knowledge} gap by incorporating retrieved node documentation;
(2) reinforcement learning (RL) for incentivizing \textit{reasoning} capability in LLMs.
Unlike math tasks, which typically have a single correct answer and have been extensively studied in RL settings~\cite{li2025torlscalingtoolintegratedrl,system2_thinking}, 
the reward design of workflow generation introduces multiple layers of complexity:
the generated code should contain no hallucinated nodes, and
conform to a valid format that can form a directed acyclic graph (DAG). 
Here we propose a fine-grained rule-metric hybrid reward.
If there is anything wrong with the format validity, structural integrity, or node hallucinations, 
it will receive a negative reward to penalize such errors. 
Only when the above rules are satisfied,
indicating that the workflow is well-formed,
a positive reward will be given based on the correctness of node selection.
We empirically demonstrate the effectiveness of this reward mechanism using Group Relative Policy Optimization~\cite{shao2024deepseekmath}.

Experiments on our test set show that, building upon Qwen2.5-Coder \cite{hui2024qwen25coder}, 
our 7B ComfyUI-R1 model outperforms previous state-of-the-art methods that rely on top-tier closed-source LLMs such as GPT-4o and the Claude series. 
It achieves a high format validity rate of 97\%,
a substantial improvement over the original Qwen2.5-Coder model (which achieves only 41\%), showing the effectiveness of our two-stage training.
ComfyUI-R1 also delivers superior performance in node-level and graph-level F1 scores. 
We further evaluate end-to-end retrieval and generation performance on ComfyBench~\cite{xue2024comfybench},
where ComfyUI-R1 surpasses the previous state-of-the-art ComfyAgent by an absolute margin of 11\% in pass rate.
Additional analysis underscores the importance of the reasoning process and the advantage of using code-based workflow representations instead of JSON. 
Qualitative comparisons demonstrate \ours's ability to synthesize complex workflows with diverse nodes, showcasing the potential of long CoT reasoning in AI-driven content creation.

In summary, our main contributions are as follows:

\begin{itemize}
\item We present \ours, the first large reasoning model for automated workflow generation to the best of our knowledge, incentivizing long CoT reasoning capability for complex workflow planning.
\item We curate a comprehensive workflow and node knowledge base and develop a two-stage training framework that combines supervised fine-tuning for cold start and reinforcement learning with our proposed rule-metric hybrid reward.
\item Extensive experiments show the superior performance of the 7B ComfyUI-R1 over previous state-of-the-art methods based on GPT-4o and the Claude series, underscoring the potential of long CoT reasoning in AI art creation.
\end{itemize}

\section{Related Work}

\subsection{General Workflow Generation}

General workflow generation has seen significant advancements with the advent of AI systems, such as Deep Research~\cite{zheng2025deepresearcher,openai2025deep_research}, OpenAI o3~\cite{openai2025o3o4}, which are capable of converting natural language descriptions into structured execution plans. 
Early efforts in this domain focus on straightforward human-designed workflows that constrain the planning process of large language or multimodal models (LLMs/LMMs) to prevent hallucinations~\cite{guo2024multiagent_survey}.
Through such frameworks, multiple models can collaborate effectively and demonstrate collective intelligence in various domains, ranging from sophisticated reasoning tasks~\cite{wang2025mixtureofagents,liu2025breaking,xu2023reasoning,chen2024agentverse} and software development~\cite{hong2024metagpt,qian2023chatdev}, to long-document translation~\cite{wu2024perhapshumantranslation} and video generation~\cite{xu2025filmagent,anim-director}.
However, these manually crafted workflows are time-consuming to develop, heavily dependent on domain expertise, and often lack flexibility.

To address this challenge, recent studies have focused on training models or developing language agents to automatically generate workflows~\cite{shang2025agentsquare,niu2025flow,hu2025adas,llm4workflow}.
For example, WorkflowLLM~\cite{fan2025workflowllm} proposes a data-centric framework specifically designed to enhance the workflow orchestration capabilities of LLMs by fine-tuning on a large-scale dataset of complex, real-world workflows.
AFlow~\cite{zhang2025aflow} reformulates the problem of workflow optimization as a search task.
By employing Monte Carlo Tree Search (MCTS), AFlow efficiently explores the expansive action space, iteratively refining workflows based on tree-structured experiences and execution feedback.
WorFEval~\cite{qiao2025worfeval} is a recently released benchmark for assessing the quality of generated workflows,
underscoring the need for models capable of generating complex graph structures.
These studies collectively reflect a shift towards building systems that can manage more intricate and realistic workflow demands.
The ComfyUI platform~\cite{comfyanonymous2023comfyui} stands out as a particularly challenging testbed in this context,
due to the large number of components involved in AI art creation workflows and the intricate connections between different modules with diverse functionalities~\cite{xue2024comfybench}.

\subsection{ComfyUI-based Workflow Generation}

Recent advances in AI-generated content have transitioned from end-to-end diffusion models~\cite{NEURIPS2020_4c5bcfec_diffusion,NEURIPS2021_49ad23d1_diffusion,wang2025unifiedagentic,podell2023sdxl} 
to more sophisticated workflows on open-source WebUI platforms such as ComfyUI.
Users can easily construct workflows by connecting a series of blocks,
such as large language models (LLMs) for refining input prompts, LoRAs trained to introduce specific artistic styles, improved latent decoders for finer details, super-resolution blocks, and more~\citep{hu2021lora,manas2024texttoimage,super_resolution}.
Creating a well-designed workflow and selecting appropriate nodes requires significant expertise~\cite{comfyui_copilot},
where automated workflow generation comes into help.

ComfyUI-based workflow generation has emerged as a trending research topic in AI-driven multimodal content synthesis.
For example,
\citet{xue2024comfybench} introduce a multi-agent ComfyAgent framework to design workflows in a step-by-step manner, consisting of three independent modules: Memory, Planner, and Actions.
Similarly, ComfyGPT~\cite{huang2025comfygpt} introduces a self-optimizing system of four specialized LLM agents, 
focusing on node connection logic.
Despite these advancements, the field remains in its early stages and faces two key limitations.
Firstly, some existing approaches are limited to text-to-image generation~\cite{sobania2024comfygi,gal2024comfygen}, restricting applicability to broader multimodal tasks. 
Secondly, the generated JSON workflows often suffer from structural inconsistencies or contain incorrect, hallucinated components,
likely due to LLMs' limited understanding of newly introduced nodes and error propagation in multi-agent frameworks~\cite{huang2025comfygpt,guo2024multiagent_survey}.
To overcome these issues, our work expands the scope to multimodal tasks and explores the post-training of foundational LLMs rather than relying on proprietary APIs, 
with the goal of improving robustness and generalization in automated workflow generation.

\subsection{Large Reasoning Models.}

Since the release of OpenAI o1~\cite{openai2024introducing}, 
large reasoning models have rapidly become a focal point in AI research, simulating ``System 2'' slow thinking processes~\cite{system2_thinking,li2025perceptionreasonthinkplan}. 
Early attempts to replicate o1 rely on supervised fine-tuning (SFT) using chain-of-thought traces, augmented with tree search algorithms and self-reflection strategies~\cite{shinn2023reflexion,macro-o1,li2025search}. 
Inspired by the success of DeepSeek-R1
—which demonstrates strong performance on challenging benchmarks in mathematics, coding, and multi-hop question answering, rivaling or even surpassing human experts~\cite{deepseekai2025deepseekr1}—
recent studies have integrated the reasoning process directly into the optimization loop via group relative policy optimization (GRPO)~\cite{shao2024deepseekmath}.
This line of work includes re-implementations of R1~\cite{openr1}, extensions to multimodal settings~\cite{zhao2025r1omni,peng2025lmmr1,zhou2025VisualThinker-R1-Zero}, applications in search~\cite{song2025r1}, tool integration~\cite{qian2025toolrl,li2025torlscalingtoolintegratedrl}, and medicine and finance domains~\cite{pan2025medvlm,liu2025finr1}. 
Overall, the post-training landscape has progressed from SFT with curated rationales to sophisticated reinforcement learning (RL) pipelines.
In this work, we explore reasoning models designed specifically for automated workflow generation, propose a novel reward mechanism for RL in this task, and finally develop ComfyUI-R1.

\section{ComfyUI-R1}

ComfyUI-R1 is a large reasoning model designed for automated workflow generation.
Given a user query about task description,
based on a set of retrieved nodes,
ComfyUI-R1 first performs step-by-step reasoning to select relevant nodes and plan the workflow.
It then generates the code representation of a ComfyUI workflow.
In this section, we first introduce the construction of ComfyUI-related knowledge bases (KBs) in Sec.~\ref{sec:kb}.
Then ComfyUI-R1 is trained using a two-stage pipeline:
(1) Cold-start supervised fine-tuning (SFT) with distilled long chain-of-thought (CoT) data to adapt the model to the ComfyUI domain (Sec.~\ref{sec:sft});
(2) Reinforcement learning (RL) for incentivizing reasoning capability via a fine-grained hybrid reward (Sec.~\ref{sec:reward}) and the GRPO algorithm (Sec.~\ref{sec:grpo}).

\begin{figure}[t]
\centering
  \includegraphics[width=\linewidth]{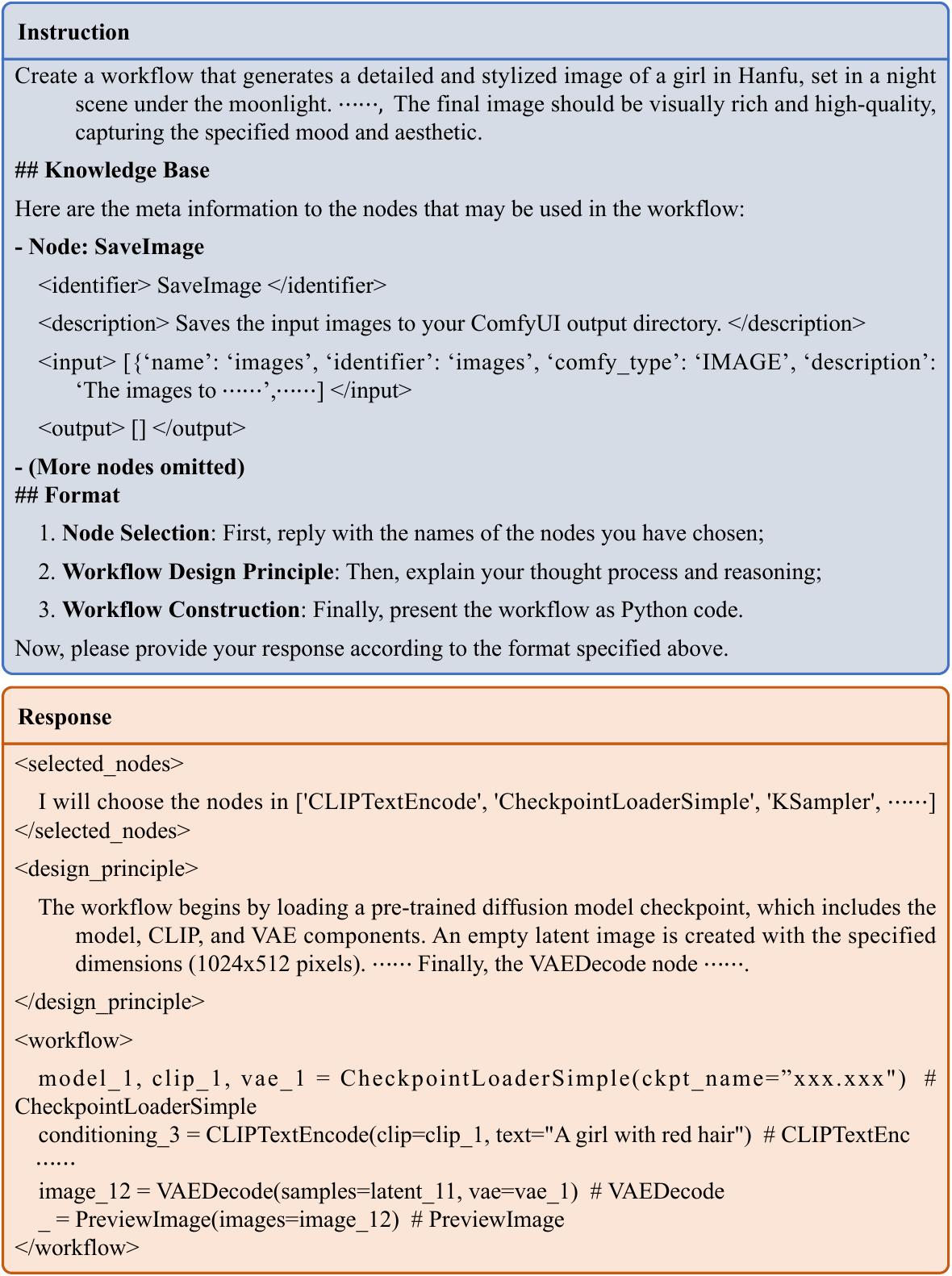}
  \caption{An example of SFT data.}
  \label{fig:sft_data}
\end{figure}

\subsection{Knowledge Bases}
\label{sec:kb}
We begin by constructing KBs of ComfyUI workflows and nodes.
The original data is sourced from popular platforms for sharing generative resources, ComfyUI-related GitHub repositories, and the ComfyUI website, with NSFW content filtered out. 
Since the raw data contains a large amount of noise,
we conduct a comprehensive cleaning process to filter and standardize the data.

\paragraph{Node KB} 
Starting with a collection of 40K nodes, we first apply exact-match deduplication. 
After filtering out nodes that lack input or output parameters, 7,238 nodes remain.
For nodes without structured documentation,
we use Claude 3.5 to generate detailed documentation by analyzing their GitHub repositories.
The final documentation is standardized to include the node type, usage, and the meanings of input and output parameters.

\paragraph{Workflow KB}
Workflows in ComfyUI are natively represented on the canvas in JSON format.
We implement a parser to extract a Directed Acyclic Graph (DAG) from the JSON file, which is then converted into a sequence of Python-like function calls in topological order.
Another parser is implemented to reverse the conversion,
ensuring that workflows can be expressed through code and remain fully compatible with the ComfyUI environment.

Starting with the collected 27K workflows, we perform a series of filtering and refinement steps.
First, we discard workflows that do not meet ComfyUI's execution standards.
Second, we remove exact duplicates.
Third, we verify whether each workflow can be successfully transformed between JSON and code representations, retaining only those that support both conversions.
Next, we further refine workflows by removing irrelevant information, such as ``Anything Anywhere'' node.
Finally, we discard workflows that contain nodes not present in the above node KB.
After this cleaning process, the resulting workflow KB contains 3,917 workflows, with an average of 21 nodes per workflow.

\begin{figure*}[t]
\centering
  \includegraphics[width=0.9\linewidth]{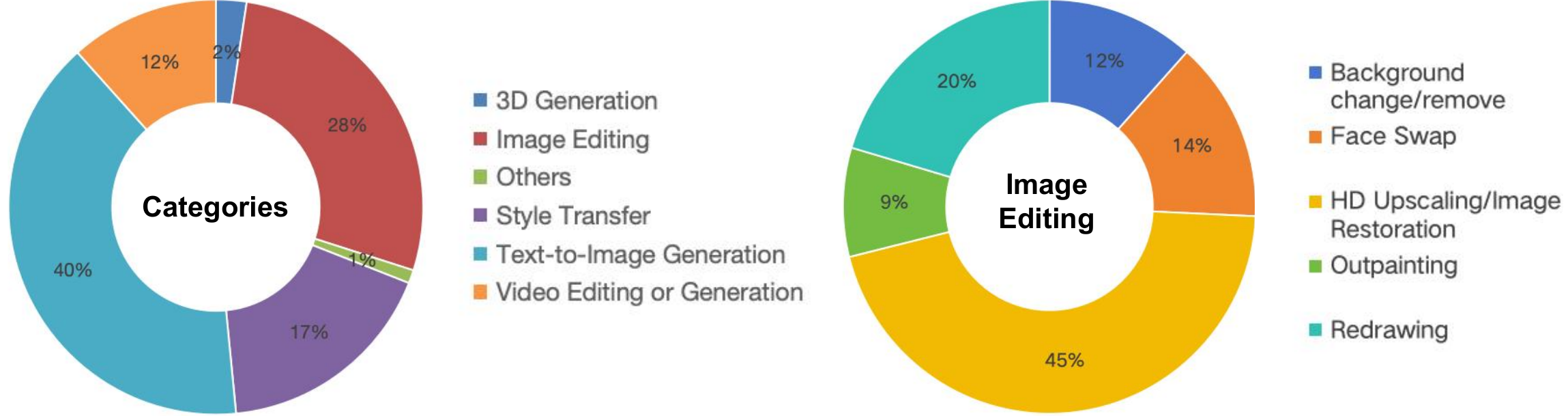}
  \caption{An illustration of the task categories in our constructed workflow KB, including the subcategories under image editing.}
  \label{fig:pie}
\end{figure*}

Since community-sourced content tends to focus more on installation instructions, there is often a lack of detailed functional descriptions for workflows. 
To address this, we leverage the multimodal understanding capabilities of GPT-4o,
by prompting it with community-sourced texts and accompanying images that typically illustrate workflow outcomes.
This approach helps fill in the gaps in informative functional descriptions.
Therefore, each entry in our workflow KB includes a JSON-formatted and code-based representation of the workflow, along with an explanation of its functionality.
As shown in Figure~\ref{fig:pie}, our workflow KB encompasses a diverse set of tasks, including text-to-image generation, image editing, style transfer, video editing and generation.

\subsection{Supervised Fine-tuning}
\label{sec:sft}

We generate long CoT reasoning sequences through a simulated node
retrieval process, node selection, and workflow planning, ultimately
producing a code representation of the workflow.
Each entry in the workflow KB is represented as a pair $(desc, c)$,
where $desc$ is the workflow description and $c$ is its corresponding code representation.
Let $\mathcal{V}^g$ denote the set of nodes in the ground-truth workflow,
and $\mathcal{V}^{\text{KB}}$ denote the full set of nodes in the KB.
To simulate the node retrieval process, 
we construct a candidate node set $\mathcal{V}^{\text{cand}}$ by 
combining $\mathcal{V}^g$ with a randomly sampled set of nodes $\mathcal{V}^{\text{random}}$,
where $\mathcal{V}^{\text{random}} \subset \mathcal{V}^{\text{KB}} \setminus \mathcal{V}^g$, and:
$$
\mathcal{V}^{\text{cand}} = \mathcal{V}^g \cup \mathcal{V}^{\text{random}}, \, \text{with } |\mathcal{V}^{\text{random}}| = \lfloor 0.8 \cdot |\mathcal{V}^g| \rfloor.
$$

Based on the workflow description $desc$ and its corresponding code representation $c$,
we generate user instructions $query$ and a rationale $r$ about the workflow design principle
using Qwen-Max, Claude 3.5 and GPT-4o.
We split 3,717 workflows into training and 200 for testing.
This results in 11,140 training samples and 600 testing samples.
As shown in Figure~\ref{fig:sft_data},
each sample consists of the user instruction $query$, 
a set of candidate nodes $\mathcal{V}^{\text{cand}}$, 
the workflow planning rationale $r$, 
the nodes in the gold workflow $\mathcal{V}^g$, 
and the final code representation $c$.

During the SFT phase, 
the model input includes the $query$ and the candidate nodes $\mathcal{V}^{\text{cand}}$.
The output is a long CoT reasoning sequence, denoted as $s=[\mathcal{V}^g,r,c]$.
The training objective can be formulated as:
\begin{equation}
    \mathcal{L}_{\text{SFT}}
  = -
      \log \sum_{t=1}^{T}
        \text{Pr} \, \bigl(s_{,t} = i \,\big|\, \text{desc}, \mathcal{V}^{\text{cand}},\, s_{<t}\bigr)
\end{equation}
where $s_{<t}$ is the sequence of tokens generated before time step $t$,
$\text{Pr} \, \bigl(s_{,t} = i \,\big|\, \text{desc}, \mathcal{V}^{\text{cand}},\, s_{<t}\bigr)$ is the probability that the LLM predicted token $s_{,t}$ in step $t$.

\subsection{Reward Design}
\label{sec:reward}

After adapting LLMs to the ComfyUI domain via SFT,
we conduct RL to further enhance the reasoning capability.
For an RL algorithm to be effective,
a well-designed \textbf{reward} is essential.
Unlike mathematical tasks in prior work—which typically have a single correct answer and allow for simple rule-based rewards—the workflow generation task presents several layers of complexity.
For example, the generated code should avoid hallucinated
nodes, and adhere to a valid format that forms a directed acyclic graph (DAG).
To this end, here we propose a fine-grained rule-metric hybrid reward $R_{\text{final}}$, including format reward $R_{\text{format}}$, structure reward $R_{\text{DAG}}$, node fidelity $R_{\text{fidelity}}$ and precision reward $R_{\text{correct}}$.
This reward formulation penalizes invalid formats, incorrect graph structures, and hallucinated nodes, while promoting accurate node selection.


\paragraph{Format reward ($R_{\text{format}}$).} 
This component verifies whether the output response adheres to the expected structure, including a sequence of reasoning steps enclosed within the 
``<selected\_nodes>...</selected\_ nodes>'' and 
``<design\_principle>...</design\_principle>'' tags.
The final workflow code is wrapped within ``<workflow>...</workflow>''.
A format score of 0 is given if all required tags are present and their contents can be successfully extracted. Otherwise, a penalty is applied.
The format reward function $R_{\text{format}}$ is defined as follows:
      \begin{equation}
          R_{\text{format}} = \begin{cases} 0 & \text{if all required fields appear} \\ -1 & \text{otherwise} \end{cases}
      \end{equation}

\paragraph{Structure reward ($R_{\text{DAG}}$).} 
This reward assesses whether the parsed ``<workflow>'' section of the generated output forms a valid DAG, which is a fundamental requirement for ComfyUI workflows.
A penalty is assigned if the structure is not a valid DAG.
      \begin{equation}
          R_{\text{DAG}} = \begin{cases} 0 & \text{if the structure is a valid DAG} \\ -1 & \text{otherwise} \end{cases}
      \end{equation}

\paragraph{Node fidelity reward ($R_{\text{fidelity}}$).} 
This component penalizes the inclusion of hallucinated or inconsistent nodes during node selection and workflow generation. 
A penalty of -1 is applied in either of the following cases:
      \begin{enumerate}
          \item \textbf{Invalid Nodes}: Any node listed in the predicted ``<selected\_ nodes>'' block $\mathcal{V}^{p}$ is not present in the provided candidate nodes $\mathcal{V}^{\text{cand}}$.
          \item \textbf{Inconsistent Nodes}: The set of nodes listed in ``<selected\_ nodes>'' block does not exactly match the set of nodes parsed from the `<workflow>'' block.
      \end{enumerate}
      \begin{equation}
          R_{\text{fidelity}} = \begin{cases} -1 & \text{if invalid or inconsistent nodes are detected} \\ 0 & \text{otherwise} \end{cases}
      \end{equation}


      \paragraph{Node Selection Accuracy ($R_{\text{correct}}$).} 
      This reward measures the overlap between generated node set $\mathcal{V}^{p}$ and the ground truth node set $\mathcal{V}^{g}$, reflecting the model’s ability to select the correct nodes for the workflow. 
      \begin{equation}
          R_{\text{correct}} = \frac{|\mathcal{V}^{p} \cap \mathcal{V}^{g}|}{|\mathcal{V}^{g}|} - 1
      \end{equation}

\paragraph{Combining the above rewards}
The total reward $R_{\text{final}}$ aggregates all the individual reward components using a veto-based mechanism.
If any of $R_{\text{format}}$, $R_{\text{DAG}}$, or $R_{\text{fidelity}}$ is -1, indicating a fundamental error in format validity, structural
integrity, or node fidelity, the total reward is immediately set to -1, preventing any positive reward for an invalid output. 
Otherwise, the total reward is computed based on the node selection accuracy.
\begin{equation}
        R_{\text{final}} = \begin{cases} -1 & \text{if } R_{\text{format}} = -1 \text{ or } R_{\text{DAG}} = -1 \text{ or } R_{\text{fidelity}} = -1 \\ \frac{4 + R_{\text{correct}}}{4.0} & \text{otherwise} \end{cases}
    \end{equation}

  \subsection{RL Training with GRPO}
  \label{sec:grpo}

Based on the above hybrid reward, 
ComfyUI-R1 is trained using the Group Relative Policy Optimization (GRPO) algorithm~\citep{shao2024deepseekmath}. 
At each training iteration, given an input text \textit{q} (including the $query$ and the candidate nodes $\mathcal{V}^{\text{cand}}$), 
a group of $G$ candidate outputs, $\{s_1, s_2, \dots, s_G\}$, is sampled from the policy model $\pi_{\theta_{\text{old}}}$. 
The advantage is computed based on the group of hybrid rewards $\{r_1, r_2, \dots, r_G\}$, defined as
\begin{equation}
A_i = \frac{r_i - \operatorname{mean}(\{r_1, r_2, \dots, r_G\})}{\operatorname{std}(\{r_1, r_2, \dots, r_G\})},
\label{eq:advantage}
\end{equation}
The GRPO algorithm then optimizes the policy $\pi_{\theta}$ by maximizing the following objective function:
\begin{equation}
\begin{aligned}
J_{\mathrm{GRPO}}(\theta) 
&= \mathbb{E}_{q \sim P(Q),\, \{s_i\}_{i=1}^G \sim \pi_{\theta_{\mathrm{old}}}(O \mid q)} \\
&\!\!\!\Biggl[
  \frac{1}{G} \sum_{i=1}^G
  \min\,\Bigl(
    \frac{\pi_{\theta}(s_i \mid q)}{\pi_{\theta_{\mathrm{old}}}(s_i \mid q)}\,A_i,\, \\
    &\!\!\!\mathrm{clip}\,\Bigl(
      \frac{\pi_{\theta}(s_i \mid q)}{\pi_{\theta_{\mathrm{old}}}(s_i \mid q)},
      1-\varepsilon,\,
      1+\varepsilon
    \Bigr)
    A_i
  \Bigr)  \\
  &\!\!\!-\,\beta\,D_{\mathrm{KL}}\bigl(\pi_{\theta}\,\big\|\,\pi_{\mathrm{ref}}\bigr)
\Biggr]
\end{aligned}
\label{eq:rl_loss}
\end{equation},
\begin{equation}
\mathrm{D}_{\mathrm{KL}}\!\bigl(\pi_\theta \,\|\, \pi_{\mathrm{ref}}\bigr)
  = \frac{\pi_{\mathrm{ref}}(o_i \mid q)}{\pi_\theta(o_i \mid q)}
    \;-\;
    \log\!\frac{\pi_{\mathrm{ref}}(o_i \mid q)}{\pi_\theta(o_i \mid q)}
    \;-\; 1,
\end{equation}
where $\varepsilon$ specifies clipping range, and $\beta$ is a hyperparameter controlling the weight of the Kullback–Leibler (KL) divergence penalty.

\section{Experiments}

\begin{table*}
  \caption{\textbf{Evaluation results of all baselines on our test set.} The node-level and graph-level \textit{precision}, \textit{recall} and \textit{F1} scores, together with \textit{Format Validity} rate, are reported. The best results are highlighted in \textbf{bold}.}
  \label{tab:main_result}
  \begin{tabular}{llccccccc}
    \toprule
    \multirow{2}{*}{\textbf{Methods}} & \multirow{2}{*}{\textbf{Models}} & \multirow{2}{*}{\textbf{Format Validity}} & \multicolumn{3}{c}{\textbf{Node-level}} & \multicolumn{3}{c}{\textbf{Graph-level}} \\
    \cmidrule(r){4-6}  \cmidrule(r){7-9}
    & & & \multicolumn{1}{c}{P} & \multicolumn{1}{c}{R} & \multicolumn{1}{c}{F1} & \multicolumn{1}{c}{P} & \multicolumn{1}{c}{R} & \multicolumn{1}{c}{F1} \\
    \midrule
   \multirow{5}{*}{Few-shot} 
    &  Qwen2.5-Coder-7B-Instruct  & 0.25 & 0.18 & 0.12 & 0.14 & 0.08 & 0.07 & 0.08 \\
    &  Qwen2.5-Max                & 0.50 & 0.35 & 0.25 & 0.29 & 0.19 & 0.17 & 0.18 \\
    &  GPT-4o                     & 0.89 & 0.62 & 0.42 & 0.50 & 0.33 & 0.28 & 0.30 \\
    &  Claude 3.5 Sonnet          & 0.93 & 0.58 & 0.48 & 0.52 & 0.37 & 0.36 & 0.37 \\
    &  Claude 3.7 Sonnet          & 0.81 & 0.44 & 0.43 & 0.43 & 0.27 & 0.29 & 0.28 \\
    \midrule
   \multirow{5}{*}{CoT} 
    &  Qwen2.5-Coder-7B-Instruct  & 0.41 & 0.30 & 0.18 & 0.22 & 0.12 & 0.09 & 0.10 \\
    &  Qwen2.5-Max                & 0.64 & 0.46 & 0.31 & 0.36 & 0.25 & 0.21 & 0.23 \\
    &  GPT-4o                     & 0.92 & 0.66 & 0.42 & 0.50 & 0.33 & 0.27 & 0.29 \\
    &  Claude 3.5 Sonnet          & \textbf{0.97} & \textbf{0.70} & 0.49 & 0.57 & 0.41 & 0.36 & 0.38 \\
    &  Claude 3.7 Sonnet          & 0.90 & 0.57 & 0.48 & 0.51 & 0.36 & 0.35 & 0.35 \\
    \midrule
    ComfyAgent & GPT-4o  & 0.47 & 0.26 & 0.20 & 0.21 & 0.11 & 0.10 & 0.10 \\
    \midrule
     {\textbf{SFT + GRPO}}& {\textbf{ComfyUI-R1}} & \textbf{0.97} & 0.67 & \textbf{0.58} & \textbf{0.62} & \textbf{0.52} & \textbf{0.51} & \textbf{0.51} \\
    \bottomrule
\end{tabular}
\end{table*}

\subsection{Data and Evaluation Metric}
Our experiments are conducted on the test set described in Sec.~\ref{sec:sft}.
Each task input consists of a $query$ and a set of candidate nodes $\mathcal{V}^{\text{cand}}$,
and the expected output is an executable workflow $c$ that addresses the given query.
Our evaluation setup is designed to assess the model’s ability to reason over pre-retrieved node information, rather than its retrieval capability. 
To explore end-to-end performance—including both node retrieval and workflow generation—we conduct an additional experiment in Sec.~\ref{sec:comfybench}, where no candidate set $\mathcal{V}^{\text{cand}}$ is provided.

The first evaluation metric is \textbf{Format Validity} rate.
It provides an initial check of the syntactic and structural correctness of the generated workflows. Specifically, it verifies whether all node names referenced in the workflow exist and whether the resulting structure forms a valid DAG.

Following the evaluation protocol in WorFEval~\cite{qiao2025worfeval},
we quantitatively assess the matching of the predicted and gold workflows.
Specifically, we identify the longest node chain via Longest Increasing Subsequence (LIS)\footnote{\url{https://en.wikipedia.org/wiki/Longest_increasing_subsequence}}
and the largest workflow subgraph via Maximum Common Induced Subgraph (MCIS)\footnote{\url{https://en.wikipedia.org/wiki/Maximum_common_induced_subgraph}}.
Based on these matches,
we compute and report
\textbf{node-level and graph-level precision, recall and F1 scores}\footnote{For detailed derivations and formal equations of the evaluation metrics, please refer to the original WorFEval paper. Our implementation follows its official GitHub repository.}.

\subsection{Baselines}
Following previous workflow generation studies~\cite{huang2025comfygpt,xue2024comfybench},
we adopt two effective prompting methods for advanced commercial LLMs (Qwen-Max, Claude 3.5, Claude 3.7 and GPT-4o),
as well as Qwen2.5-Coder-7B-Instruct, which serves as the base model for our training.
\begin{enumerate}
    \item \textbf{Few-shot learning} provides a set of code-formatted workflows in the prompt to utilize the in-context learning capabilities of LLMs for workflow generation.
    \item \textbf{Chain-of-Thought (CoT)} is improved based on the above few-shot learning baseline, incorporating the CoT reasoning process into the few-shot demonstrations.
\end{enumerate}
In addition, we include the multi-agent baseline \textbf{ComfyAgent}~\cite{xue2024comfybench} based on its official GitHub implementation.

\begin{table}
  \caption{\textbf{Ablation results of ComfyUI-R1} with different settings.
  ``SFT + GRPO'' indicates ComfyUI-R1. Two variants only conduct the SFT stage, either with code or JSON-formatted workflows. ``FV'' means format validity.}
  \label{tab:ablation}
  \begin{tabular}{llccccccc}
    \toprule
    \multirow{2}{*}{\textbf{Methods}} & \multirow{2}{*}{\textbf{FV}} & \multicolumn{3}{c}{\textbf{Node-level}} & \multicolumn{3}{c}{\textbf{Graph-level}} \\
    \cmidrule(r){3-5}  \cmidrule(r){6-8}
    & & \multicolumn{1}{c}{P} & \multicolumn{1}{c}{R} & \multicolumn{1}{c}{F1} & \multicolumn{1}{c}{P} & \multicolumn{1}{c}{R} & \multicolumn{1}{c}{F1} \\
    \midrule
    SFT + GRPO      & 0.97 & 0.67 & 0.58 & 0.62 & 0.52 & 0.51 & 0.51 \\
    SFT only        & 0.95 & 0.64 & 0.57 & 0.60 & 0.48 & 0.49 & 0.48 \\
    SFT only (JSON) & 0.92 & 0.62 & 0.55 & 0.57 & 0.45 & 0.46 & 0.45 \\
    \bottomrule
\end{tabular}
\end{table}

\begin{figure*}[t]
    \centering
    \includegraphics[width=0.82\textwidth]{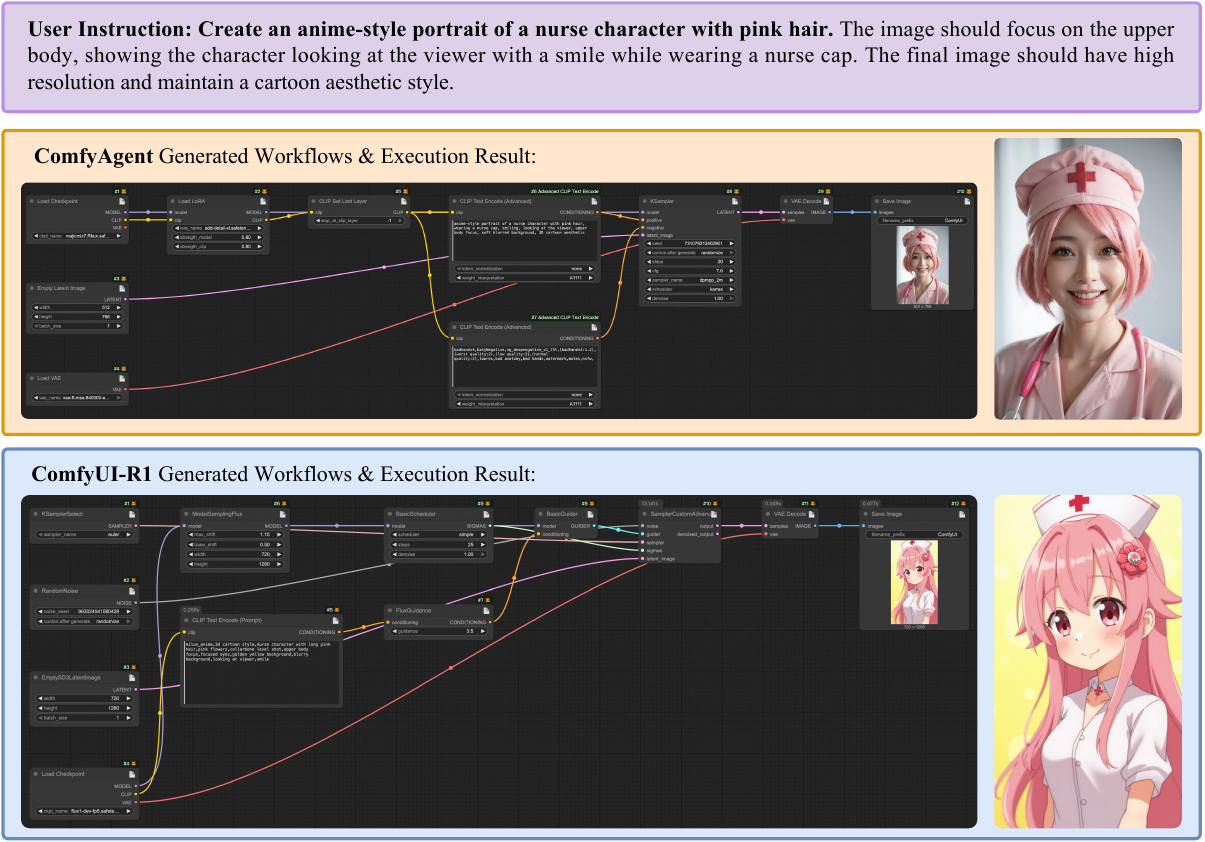}
    \caption{\textbf{Comparison between ComfyAgent and our ComfyUI-R1.} The execution result of ComfyUI-R1 accurately adheres to the ``anime-style'' and ``cartoon'' attributes in the user instruction. In contrast, ComfyAgent fails to follow these stylistic guidelines.}
    \label{fig:case}
\end{figure*}

\subsection{Implementation Details}
We use the Qwen2.5-Coder-7B-Instruct as the backbone of ComfyUI-R1.
The SFT training stage is conducted on 8$\times$80G NVIDIA A100 GPUs over 1 epoch,
with a learning rate of 1e-5 and a batch size of 1 on each GPU.
During RL training, 
we utilize 8$\times$80G NVIDIA A100 GPUs to train for a total of 300 steps,
with a learning rate of 1e-6 and a total batch size of 64.
In Eq.~\ref{eq:rl_loss},
the number of group computations $G$ is set to 4,
the clipping coefficient $\varepsilon$ to 0.2,
the KL penalty weight $\beta$ to 0.001.
The maximum tokens are set to 32,768 for both training and inference.

\subsection{Results}

Table~\ref{tab:main_result} presents the experimental results on our test set.
In terms of node-level and graph-level F1 scores,
our 7B ComfyUI-R1 model outperforms few-shot, CoT prompting and multi-agent methods, which all rely on top-tier closed-source LLMs such as GPT-4o and the Claude series. 
Compared to the original Qwen2.5-Coder model, ComfyUI-R1 achieves substantial improvements across all metrics (e.g., format validity rate increases from 41\% to 97\%), demonstrating the effectiveness of our two-stage training strategy.
When comparing performance under few-shot and CoT settings, all models show consistent improvement with CoT reasoning in their in-context demonstrations, highlighting the importance of reasoning process.

Table~\ref{tab:ablation} reports results from two variants of ComfyUI-R1:
(1) an SFT-only version, and
(2) a version that replaces code-based workflow representations with a JSON format proposed by \citet{huang2025comfygpt}.
The results show that RL training further improves the already high 95\% format validity rate of the SFT-only model, validating the effectiveness of RL in the workflow generation task.
On the other hand, 
switching from code to JSON for workflow representation leads to decreased performance, 
likely because the JSON format carries less semantic and logical structure. 
Code-based representations are more compact and semantically rich, highlighting their advantage for this task.

\subsection{Additional Experiments on ComfyBench}
\label{sec:comfybench}

To evaluate the end-to-end retrieval and generation performance,
we conduct experiments on ComfyBench~\cite{xue2024comfybench} without providing candidate nodes $\mathcal{V}^{\text{cand}}$.
We report the \textit{pass rate}, which measures the ratio of tasks where generated workflows can be successfully executed in the ComfyUI server.

\begin{table}
  \caption{\textbf{Evaluation results on ComfyBench.} The numbers of GPT-4o methods are taken from the original paper~\cite{xue2024comfybench}.}
  \label{tab:comfybench}
  \begin{tabular}{lc}
    \toprule
    \textbf{Methods} & \textbf{Pass Rate} \\
    \midrule
    GPT-4o + Few-shot & 0.23 \\
    GPT-4o + CoT & 0.28 \\
    GPT-4o + RAG & 0.52 \\
    GPT-4o + ComfyAgent & 0.56 \\
    \midrule
    ComfyUI-R1 & \textbf{0.67} \\
    \bottomrule
\end{tabular}
\end{table}

Our retrieval process for obtaining candidate nodes $\mathcal{V}^{\text{cand}}$ is as follows:
(1) Given an instruction, we retrieve the top 3 most semantically relevant workflows from the workflow KB based on OpenAI's \texttt{text-embedding-3-small}.
(2) We then aggregate the nodes from these retrieved workflows into a set, which forms $\mathcal{V}^{\text{cand}}$.
Next, the $query$ and candidate nodes are input into ComfyUI-R1 for workflow generation.
As shown in Table~\ref{tab:comfybench},
our model outperforms previous state-of-the-art, the GPT-4o-based ComfyAgent, by an absolute margin of 11\%, 
showing the effectiveness of our training scheme tailored for the workflow generation task.
Sec.~\ref{sec:cases} provides more qualitative comparison of ComfyAgent and our ComfyUI-R1.

\subsection{Case Study}
\label{sec:cases}

\begin{figure*}[t]
    \centering
    \includegraphics[width=0.75\textwidth]{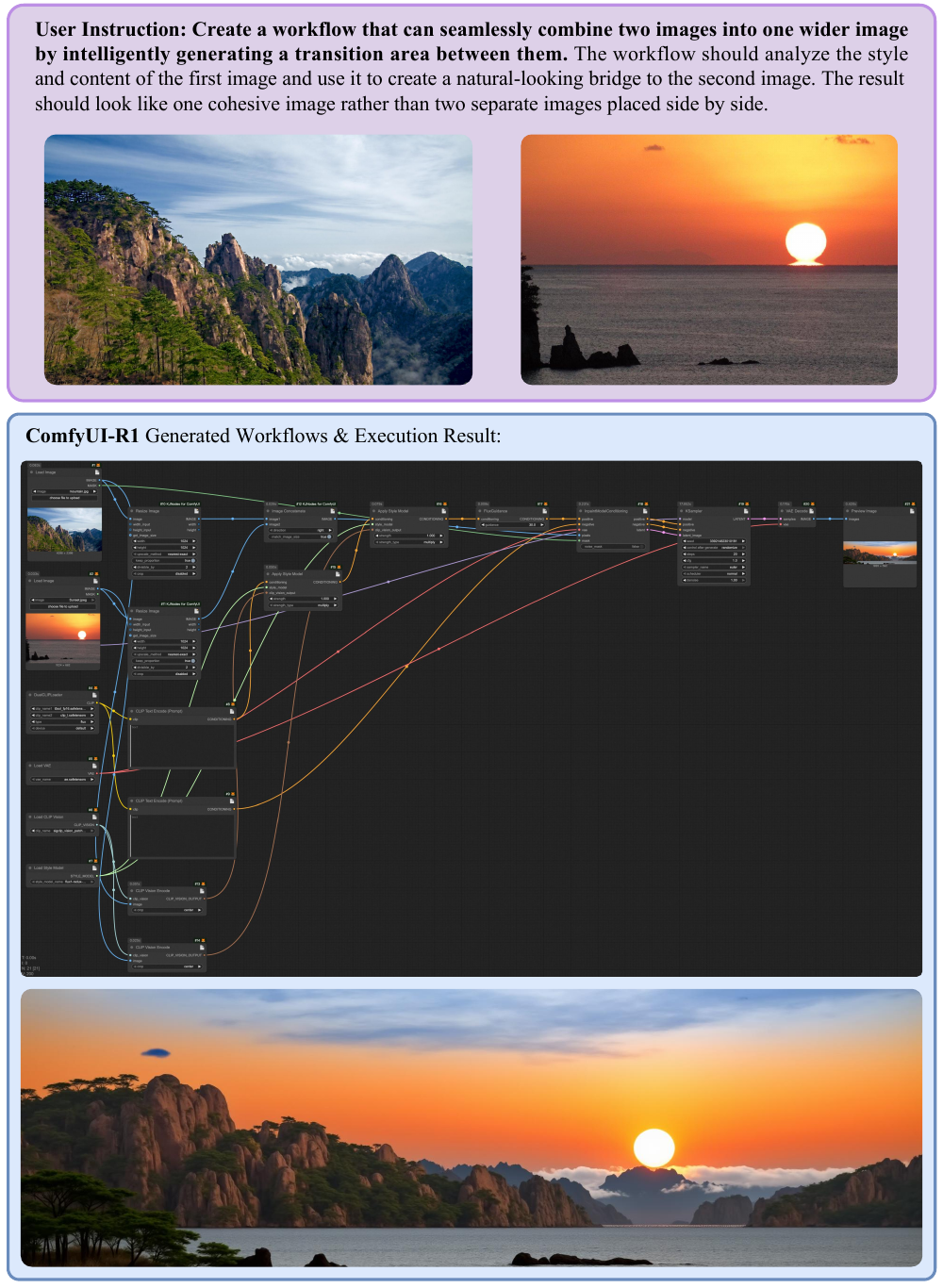}
    \caption{\textbf{Case study: multi-image combination.} Here ComfyUI-R1 seamlessly combines the input images by generating a correct and executable workflow.}
    \label{fig:appen_r1}
\end{figure*}

\begin{figure*}[t]
    \centering
    \includegraphics[width=0.75\textwidth]{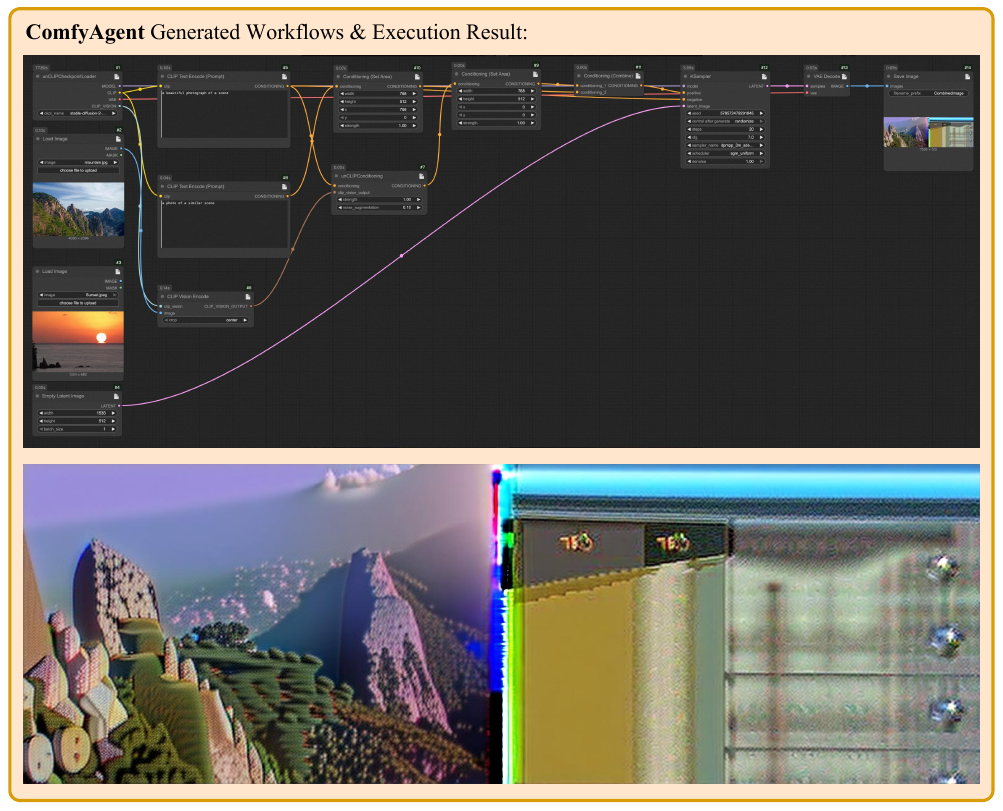}
    \caption{\textbf{Case study: multi-image combination (continued).} This figure shows the workflow and the corresponding execution result generated by ComfyAgent~\cite{xue2024comfybench}.}
    \label{fig:appen_agent}
\end{figure*}

We present a comparison between ComfyAgent and our ComfyUI-R1.
In Figure~\ref{fig:case}, the image style generated by ComfyUI-R1’s workflow aligns more closely with the user instruction compared to that produced by ComfyAgent.
From the multi-image combination example in Figures~\ref{fig:appen_r1} and~\ref{fig:appen_agent}, 
ComfyUI-R1 successfully loads and seamlessly blends the two input images. 
In contrast, ComfyAgent's workflow loads the second image but fails to utilize it further, resulting in an incomplete output. 
This highlights the importance of effective workflow planning.
From these examples, we observe that ComfyUI-R1’s workflows typically include more nodes than those of ComfyAgent, 
demonstrating \ours's capability to synthesize complex, executable, and instruction-aligned workflows with diverse nodes.


\section{Conclusion}

We introduce \textbf{ComfyUI-R1}, the first large reasoning model for automated workflow generation on the ComfyUI platform.
We begin by curating workflow and node knowledge bases to construct long chain-of-thought reasoning data.
Following this, we employ supervised fine-tuning for cold start and reinforcement learning for incentivizing reasoning capability,
guiding the model to generate structurally sound and executable workflows.
Experimental results demonstrate that our 7B ComfyUI-R1 model significantly outperforms existing SOTA methods based on GPT-4o and Claude series, achieving a 97\% format validity rate and superior node-level and graph-level F1 scores.
Qualitative analysis highlights the model’s ability to synthesize complex workflows, showcasing the power of structured reasoning in AI-driven content creation.
Future directions include exploring more fine-grained reward signals during training to better guide the intricate reasoning required for workflow generation.

\bibliographystyle{ACM-Reference-Format}
\bibliography{0_main}


\end{document}